%% file: main.tex
\documentclass{article} 
\usepackage{iclr2021_conference,times}

\input{math_commands.tex}

\usepackage[utf8]{inputenc} 
\usepackage[T1]{fontenc}    
\usepackage{hyperref}       
\usepackage{url}            
\usepackage{booktabs, arydshln}       
\usepackage{amsfonts}       
\usepackage{nicefrac}       
\usepackage{microtype}      
\usepackage{tabularx}

\usepackage{xcolor}

\usepackage{subcaption}
\usepackage{tikz}
\usetikzlibrary{arrows,positioning,shapes,quotes, patterns,shapes.symbols,shapes.callouts}
\usetikzlibrary{arrows.meta}

\usepackage{caption}
\usepackage{subcaption}
\usepackage[width=.9\textwidth]{caption}
\usepackage{xcolor}
\usepackage{array}

\usepackage{amsthm}
\usepackage{amsmath}
\usepackage{amsfonts}
\usepackage{amssymb}
\usepackage{mathtools}
\usepackage{esvect}
\usepackage{adjustbox}

\usepackage{graphicx}
\iclrfinalcopy

\makeatletter
\def\adl@drawiv#1#2#3{%
        \hskip.5\tabcolsep
        \xleaders#3{#2.5\@tempdimb #1{1}#2.5\@tempdimb}%
                #2\z@ plus1fil minus1fil\relax
        \hskip.5\tabcolsep}
\newcommand{\cdashlinelr}[1]{%
  \noalign{\vskip\aboverulesep
           \global\let\@dashdrawstore\adl@draw
           \global\let\adl@draw\adl@drawiv}
  \cdashline{#1}
  \noalign{\global\let\adl@draw\@dashdrawstore
           \vskip\belowrulesep}}
\makeatother

\newcommand{\Eb}[2]{\E_{#1}\left[#2\right]}

\newcommand{\bb}[1]{\mathbf{#1}}
\newcommand{\bbb}{\bb{b}}
\newcommand{\ba}{\bb{a}}
\newcommand{\bx}{\bb{x}}

\newcommand{\bt}{\bb{t}}

\newcommand{\bv}{\bb{v}}
\newcommand{\bw}{\bb{w}}

\newcommand{\bg}{\bb{g}}

\newcommand{\bu}{\bb{u}}

\newcommand{\bz}{\bb{z}}

\newcommand{\bbf}{\bb{f}}

\usepackage{color}
\definecolor{light}{rgb}{0.5, 0.5, 0.5}

\DeclareMathAlphabet\mathbfcal{OMS}{cmsy}{b}{n}

\newcommand{\bT}{\boldsymbol{\theta}}

\newcommand{\balpha}{\boldsymbol{\alpha}}
\newcommand{\bphi}{\boldsymbol{\phi}}

\newcommand{\pT}{p_{\bT}}
\newcommand{\PT}{P_{\bT}}
\newcommand{\pz}{p_{\bz}}

\newcommand{\qPhi}{q_{\bphi}}

\newcommand{\fT}{\bbf_{\bT}}

\newcommand{\loss}{\mathcal{L}}

\newcommand{\KLD}[2]{D_{\mathrm{KL}} \left( \left. \left. #1 \right|\right| #2 \right) }

\title{Reducing the Computational Cost of Deep \\ Generative Models with Binary Neural \\ Networks}

\author{%
 Thomas Bird*, Friso H. Kingma\textsuperscript{\textdagger} \& David Barber* \\
  *~Department of Computer Science, University College London \\
  \texttt{<firstname>.<surname>@cs.ucl.ac.uk} \\
  \textsuperscript{\textdagger} \texttt{fhkingma@gmail.com}

}

\begin{document}

\maketitle

\begin{abstract}
Deep generative models provide a powerful set of tools to understand real-world data. But as these models improve, they increase in size and complexity, so their computational cost in memory and execution time grows. Using binary weights in neural networks is one method which has shown promise in reducing this cost. However, whether binary neural networks can be used in generative models is an open problem. In this work we show, for the first time, that we can successfully train generative models which utilize binary neural networks. This reduces the computational cost of the models massively. We develop a new class of binary weight normalization, and provide insights for architecture designs of these binarized generative models. We demonstrate that two state-of-the-art deep generative models, the ResNet VAE and Flow++ models, can be binarized effectively using these techniques. We train binary models that achieve loss values close to those of the regular models but are 90\%-94\% smaller in size, and also allow significant speed-ups in execution time. 
\end{abstract}

\section{Introduction}
As machine learning models continue to grow in number of parameters, there is a corresponding effort to try and reduce the ever-increasing memory and computational requirements that these models incur. One method to make models more efficient is to use neural networks with weights and possibly activations restricted to be binary-valued \citep{binaryconnect, bnn_bengio, xnornet, wideres, projection}. Binary weights and activations require significantly less memory, and also admit faster low-level implementations of key operations such as linear transformations than when using the usual floating-point precision.

Although the application of binary neural networks for classification is relatively well-studied, there has been no research that we are aware of that has examined whether binary neural networks can be used effectively in \textit{unsupervised} learning problems. Indeed, many of the deep generative models that are popular for unsupervised learning do have high parameter counts and are computationally expensive \citep{attention, biva, flowpp}. These models would stand to benefit significantly from converting the weights and activations to binary values, which we call \textit{binarization} for brevity.

In this work we focus on non-autoregressive models with explicit densities. One such class of density model is the variational autoencoder (VAE) \citep{aevb, dlgm}, a latent variable model which has been used to model many high-dimensional data domains accurately. The state-of-the-art VAE models tend to have deep hierarchies of latent layers, and have demonstrated good performance relative to comparable modelling approaches \citep{blei_hierarchical, iaf, biva}. Whilst this deep hierarchy makes the model powerful, the model size and compute requirements increases with the number of latent layers, making very deep models resource intensive.

Another class of density model which has shown promising results are flow-based generative models \citep{nice, normalizing_flows, realnvp}. These models perform a series of invertible transformations to a simple density, with the transformed density approximating the data-generating distribution. Flow models which achieve state-of-the-art performance compose many transformations to give flexibility to the learned density \citep{glow, flowpp}. Again the model computational cost increases as the number of transformations increases.




To examine how to binarize hierarchical VAEs and flow models successfully, we take two models which have demonstrated excellent modelling performance - the ResNet VAE \citep{iaf} and the Flow++ model \citep{flowpp} - and implement the majority of each model with binary neural networks. Using binary weights and activations reduces the computational cost, but also decreases the representational capability of the model. Therefore our aim is to strike a balance between reducing the computational cost and maintaining good modelling performance. We show that it is possible to decrease the model size drastically, and allow for significant speed ups in run time, with only a minor impact on the achieved loss value. We make the following key contributions:
\begin{itemize}
    \item We propose an efficient binary adaptation of weight normalization, a reparameterization technique often used in deep generative models to accelerate convergence. \textit{Binary weight normalization} is the generative-modelling alternative to the usual batch normalization used in binary neural networks.
    \item We show that we can binarize the majority of weights and activations in deep hierarchical VAE and flow models, without significantly hurting performance. We demonstrate the corresponding binary architecture designs for both the ResNet VAE and the Flow++ model. 
    \item We perform experiments on different levels of binarization, clearly demonstrating the trade-off between binarization and performance. 
\end{itemize}

\section{Background}

In this section we give background on the implementation and training of binary neural networks.  We also describe the generative models that we implement with binary neural networks in detail.

\subsection{Binary Neural Networks}\label{sec:bnns}

In order to reduce the memory and computational requirements of neural networks, there has been recent research into how to effectively utilise networks which use binary-valued weights $\bw_{\mathbb{B}}$ and possibly also activations $\balpha_{\mathbb{B}}$ rather than the usual real-valued\footnote{We use real-valued throughout the paper to be synonymous with "implemented with floating-point precision".} weights and activations \citep{binaryconnect, bnn_bengio, xnornet, wideres, projection}. In this work, we use the convention of binary values being in $\mathbb{B}\coloneqq \{-1, 1\}$. 

\textbf{Motivation. } The primary motivation for using binary neural networks is to decrease the memory and computational requirements of the model. Clearly binary weights require less memory to be stored: $32\times$ less than the usual 32-bit floating-point weights.

Binary neural networks also admit significant speed-ups. A reported $2\times$ speed-up can be achieved by a layer with binary weights and real-valued inputs \citep{xnornet}. This can be made an additional $29\times$ faster if the inputs to the layer are also constrained to be binary \citep{xnornet}. With both binary weights and inputs, linear operators such as convolutions can be implemented using the inexpensive XNOR and bit-count binary operations. A simple way to ensure binary inputs to a layer is to have a binary activation function before the layer \citep{bnn_bengio, xnornet}.


\textbf{Optimization.} Taking a trained model with real-valued weights and binarizing the weights has been shown to be lead to significant worsening of performance \citep{gal_review}. So instead the binary weights are optimized. It is common to not optimize the binary weights directly, but instead optimize a set of underlying real-valued weights $\bw_{\mathbb{R}}$ which can then be binarized in some fashion for inference. In this paper we will adopt the convention of binarizing the underlying weights using the sign function (see Equation \ref{w-sign}). We also use the sign function as the activation function when we use binary activations (see Equation \ref{a-sign}, where $\balpha_{\mathbb{R}}$ are the real-valued pre-activations). We define the sign function as:
\begin{equation}
    \text{sign}(x) \coloneqq
\begin{cases}
    -1,& \text{if } x < 0\\
    1, & \text{if } x \geq 0
\end{cases}
\end{equation}

Since the derivative of the sign function is zero almost everywhere\footnote{Apart from at 0, where it is non-differentiable.}, the gradients of the underlying weights $\bw_\mathbb{R}$ and through binary activations are zero almost everywhere. This makes gradient-based optimization challenging. To overcome this issue, the straight-through estimator (STE) \citep{ste} can be used. When computing the gradient of the loss $\loss$, the STE replaces the gradient of the sign function (or other discrete output functions) with an approximate surrogate. A straightforward and widely used surrogate gradient is the identity function, which we use to calculate the gradients of the real-valued weights $w_{\mathbb{R}}$ (see Equation \ref{w-gradient}). It has been shown useful to cancel the gradients when their magnitude becomes too large \citep{binaryconnect, gal_review}. Therefore we use a clipped identity function for the gradients of the pre-activations (see Equation \ref{a-gradient}). This avoids saturating a binary activation. Lastly, the loss value only depends on the sign of the real-valued weights. Therefore, the values of the weights are generally clipped to be in $[-1, 1]$ after each gradient update (see Equation \ref{w-update}). This restricts the magnitude of the weights and thus makes it easier to flip the sign.
\begin{tabular}{p{1cm}p{5.5cm}p{5cm}}
{\begin{align*}
\\
\text{\textbf{Forward pass:}}&\\
\text{\textbf{Backward pass:}}&\vphantom{ \frac{\partial \loss}{\partial w_{\mathbb{R}}}}\\
\text{\textbf{After update:}}&\\
\end{align*}}
& 
{\begin{align}
&~~\text{\textbf{Weights}}\nonumber \\
w_{\mathbb{B}} &\ = \text{sign}(w_{\mathbb{R}}) \label{w-sign} \\
\frac{\partial \loss}{\partial w_{\mathbb{R}}} &\coloneqq \frac{\partial \loss}{\partial w_{\mathbb{B}}} \label{w-gradient}\\
w_{\mathbb{R}} & \leftarrow \max(-1, \min(1, w_{\mathbb{R}})) \label{w-update}
\end{align}}
&
{\begin{align}
&~~\text{\textbf{Activations}}\nonumber \\
\alpha_{\mathbb{B}} &\ = \text{sign}(\alpha_{\mathbb{R}}) \label{a-sign}\\
\frac{\partial \loss}{\partial \alpha_{\mathbb{R}}} &\coloneqq \frac{\partial \loss}{\partial \alpha_{\mathbb{B}}} * 1_{|\alpha_{\mathbb{R}}|\leq 1} \label{a-gradient}\\
&\ - \nonumber
\end{align} }
\end{tabular}

\vspace{-1cm}

\subsection{Deep Generative Models}\label{sec:gen_models}

\textbf{Hierarchical VAEs.} The variational autoencoder \citep{aevb,dlgm} is a latent variable model for observed data $\bx$ conditioned on unobserved latent variables $\bz$. It consists of a  generative model $\pT(\bx, \bz)$ and an inference model $\qPhi(\bz|\bx)$. The generative model can be decomposed into the prior on the latent variables $\pT(\bz)$ and the likelihood of our data given the latent variables $\pT (\bx|\bz)$. The inference model is a variational approximation to the true posterior, since the true posterior is usually intractable in models of interest. Training is generally performed by maximization of the evidence lower bound (ELBO), a lower bound on the log-likelihood of the data:
\begin{equation}
    \log \pT (\bx) \geq \mathbb{E}_{\qPhi(\bz|\bx)}[\log \pT(\bx, \bz) - \log \qPhi(\bz|\bx)]
\end{equation}

To give a more expressive model, the latent space can be structured into a hierarchy of latent variables $\bz_{1:L}$. In the generative model each latent layer is conditioned on deeper latents $\pT (\bz_i | \bz_{i+1:L})$. A common problem with hierarchical VAEs is that the deeper latents can struggle to learn, often "collapsing" such that the layer posterior matches the prior: $\qPhi(\bz_i|\bz_{i+1:L}, \bx) \approx \pT (\bz_i | \bz_{i+1:L})$\footnote{We have assumed here that the inference model is factored "top-down".}. One method to help prevent posterior collapse is to use skip connections between latent layers \citep{iaf, biva}, turning the layers into residual layers \citep{resnet}. 

We focus on the ResNet VAE (RVAE) model \citep{iaf}. In this model, both the generative and inference model structure their layers as residual layers. The ResNet VAE uses a bi-directional inference structure with both a bottom-up and top-down residual channel. This is a similar structure to the BIVA model \citep{biva}, which has demonstrated state-of-the-art results for a latent variable model. We give a more detailed description of the model in Appendix \ref{app:vaes}.



\textbf{Flow models.} Flow models consist of a parameterized invertible transformation, $\bz=\fT(\bx)$, and a known density $\pz(\bz)$ usually taken to be a unit normal distribution. Given observed data $\bx$ we obtain the objective for $\bT$ by applying a change-of-variables to the log-likelihood: 
\begin{equation}\label{eqn:changeofvars}
    \log \pT(\bx) = \log \pz(\fT(\bx)) + \log \bigg\vert \det \frac{d\fT }{d\bx} \bigg\vert
\end{equation}
For training to be possible, it is required that computation of the Jacobian determinant $\det (d\fT / d\bx)$ is tractable. We therefore aim to specify of flow model $\fT$ which is sufficiently flexible to model the data distribution well, whilst also being invertible and having a tractable Jacobian determinant. One common approach is to construct $\fT$ as a composition of many simpler functions: $\fT = \bbf_1 \circ \bbf_2 \circ ... \circ \bbf_L$, with each $\bbf_i$ invertible and with tractable Jacobian. So the objective becomes:
\begin{equation}
    \log \pT(\bx) = \log \pz(\fT(\bx)) + \sum_{i=1}^L \log \bigg\vert \det \frac{d\bbf_i }{d\bbf_{i-1}} \bigg\vert
\end{equation}
There are many approaches to construct the $\bbf_i$ layers \citep{nice, normalizing_flows, realnvp, glow, flowpp}. In this work we will focus on the Flow++ model \citep{flowpp}, which has state-of-the-art results for flow models. In the Flow++ model, the $\bbf_i$ are coupling layers which partition the input into $\bx_1$ and $\bx_2$, then transform only $\bx_2$:

\begin{equation}\label{eqn:flowpp_coupling}
    \bbf_i(\bx_1) = \bx_1, ~~~ \bbf_i(\bx_2) = \sigma^{-1} \big(\text{MixLogCDF}(\bx_2; \bt(\bx_1))\big) \cdot \exp(\ba(\bx_1)) + \bbb(\bx_1)
\end{equation}

Where $\text{MixLogCDF}$ is the CDF for a mixture of logistic distributions. This is an iteration on the affine coupling layer \citep{nice, realnvp}. Note that keeping part of the input fixed ensures that the layer is invertible. To ensure that all dimensions are transformed in the composition, adjacent coupling layers will keep different parts of the input fixed, often using an alternating checkerboard or stripe pattern to choose the fixed dimensions \citep{realnvp}. The majority of parameters in this flow model come from the functions $\bt, \ba$ and $\bbb$ in the coupling layer, and in Flow++ these are parameterized as stacks of convolutional residual layers. In this work we will focus on how to binarize these functions whilst maintaining good modelling performance. We give a more detailed description of the full flow model in Appendix \ref{app:flowpp}.

\section{Binarizing Deep Generative Models}\label{sec:binary_components}

In this section we first introduce a technique to effectively binarize weight normalized layers, which are used extensively in deep generative model architectures. Afterwards, we elaborate on which components of the models we can binarize without significantly hurting performance. 

\subsection{Normalization}\label{sec:norm}

It is important to apply some kind of normalization after a binary layer. Binary weights are often large in magnitude relative to the usual real-valued weights, and can result in large outputs which can destabilize training. Previous binary neural network implementations have largely used batch normalization, which can be executed efficiently using a shift-based implementation \citep{bnn_bengio}. 

However, it is common in generative modelling to use weight normalization \citep{weightnorm} instead of batch normalization. For example, it is used in the Flow++ \citep{flowpp} and state-of-the-art hierarchical VAE models \citep{iaf, biva}.  Weight normalization factors a vector of weights $\bw_{\mathbb{R}}$ into a vector of the same dimension $\bv_{\mathbb{R}}$ and a magnitude $g$, both of which are learned. The weight vector is then expressed as:
\begin{equation}\label{eqn:wn}
    \bw_{\mathbb{R}} = \bv_{\mathbb{R}} \cdot \frac{g}{||\bv_{\mathbb{R}}||}
\end{equation}
Where $||\cdot||$ denotes the Euclidean norm. This implies that the norm of $\bw_{\mathbb{R}}$ is $g$. 

Now suppose we wish to binarize the parameters of a weight normalized layer. We are only able to binarize $\bv_{\mathbb{R}}$, since binarizing the magnitude $g$ and bias $b$ could result in large outputs of the layer. However, $g$ and $b$ do not add significant compute or memory requirements, as they are applied elementwise and are much smaller than the binary weight vector. 

Let $\bv_{\mathbb{B}} = \text{sign}(\bv_{\mathbb{R}})$ be a binarized weight vector of dimension $n$. Since every element of $\bv_{\mathbb{B}}$ is one of $\pm 1$, we know that $||\bv_{\mathbb{B}}|| = \sqrt{n}$ \footnote{$||\bv_{\mathbb{B}}|| = \sqrt{\sum_i (v_{{\mathbb{B}}, i})^2} = \sqrt{\sum_i 1} = \sqrt{n}$}. We then have:
\begin{equation}
    \bw_{\mathbb{R}} = \bv_{\mathbb{B}} \cdot \frac{g}{\sqrt{n}}
\end{equation}
We refer to this as \textit{binary weight normalization}, or BWN. Importantly, this is faster to compute than the usual weight normalization (Equation \ref{eqn:wn}), since we do not have to calculate the norm of $\bv_{\mathbb{B}}$. The binary weight normalization requires only $O(1)$ FLOPs to calculate the scaling for $\bv_{\mathbb{B}}$, whereas the regular weight normalization requires $O(n)$ FLOPs to calculate the scaling for $\bv_{\mathbb{R}}$. For a model of millions of parameters, this can be a significant speed-up. Binary weight normalization also has a more straightforward backward pass, since we do not need to take gradients of the $1/||\bv||$ term.



Furthermore, convolutions $\mathcal{F}$ and other linear transformations can be implemented using cheap binary operations when using binary weights, $\bw_{\mathbb{B}}$, as discussed in Section \ref{sec:bnns}\footnote{This applies when the inputs are real-valued or binary, but the speed-ups are far greater for binary inputs}. However, after applying binary weight normalization, the weight vector is real-valued, $\bw_{\mathbb{R}}$. Fortunately, since a convolution is a linear transformation, we can apply the normalization factor $\alpha =  g / \sqrt{n}$ either before or after applying the convolution to input $\bx$. 
\begin{equation}
    \mathcal{F}(\bx, \bv_{\mathbb{B}} \cdot \alpha) = \mathcal{F}(\bx, \bv_{\mathbb{B}}) \cdot \alpha
\end{equation}

So if we wish to utilize fast binary operations for the binary convolution layer, we need to apply binary weight normalization \textit{after} the convolution. This means that the weights are binary for the convolution operation itself. This couples the convolution operation and the weight normalization, and we refer to the overall layer as a binary weight normalized convolution, or BWN convolution. Note that the above process applies equally well to other linear transformations. We initialize BWN layers in the same manner as regular weight normalization, but give a more thorough description of alternatives in Appendix \ref{app:init}. 

\subsection{Binarizing Residual Layers}\label{sec:partial_bin}

We aim to binarize deep generative models, in which it is common to utilize residual layers extensively. Residual layers are functions with skip connections:
\begin{equation}\label{eq:res}
    \bg_{\text{res}}(\bx) = \gT(\bx) + \bx
\end{equation}

Indeed, the models we target in this work, the ResNet VAE and Flow++ models, have the majority of their parameters within residual layers. Therefore they are natural candidates for binarization, since binarizing them would result in a large decrease in the computational cost of the model. To binarize them we implement $\gT(\bx)$ in Equation \ref{eq:res} using binary weights and possibly activations.

The motivation for using residual layers is that they can be used to add more representative capability to a model without suffering from the \textit{degradation problem} \citep{resnet}. That is, residual layers can easily learn the identity function by driving the weights to zero. So, if sufficient care is taken with initialization and optimization, adding residual layers to the model should not degrade performance, helping to precondition the problem.

Degradation of performance is of particular concern when using binary layers. Binary weights and activations are both less expressive than their real-valued counterparts, and more difficult to optimize. These disadvantages of binary layers are more pronounced for generative modelling than for classification. Generative models need to be very expressive, since we wish to model complex data such as images. Optimization can also be difficult, since the likelihood of a data point is highly sensitive to the distribution statistics output by the model, and can easily diverge. This provides an additional justification for binarizing only the residual layers of a generative model. By restricting binarization to the residual layers, it decreases the chance that using binary layers harms performance.

Crucially, if we were to use a residual binary layer \textit{without} weight normalization, then the layer would not be able to learn the identity function, as the binary weights cannot be set to zero. This would remove the primary motivation to use binary residual layers. In contrast, using a binary weight normalized layer in the residual layer, the gain $g$ and bias $b$ can be set to zero to achieve the identity function. As such, we binarize the ResNet VAE and Flow++ models by implementing the residual layers using BWN layers.

\begin{figure}[t]
\vspace*{-0.5cm}
\centering
\begin{subfigure}{.25\textwidth}
\centering
\includegraphics{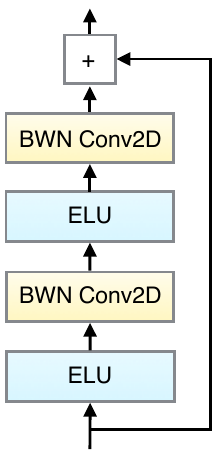}
\caption{RVAE }{(32-bit activations)}
\label{subfig:gen}
\end{subfigure}%
\begin{subfigure}{.25\textwidth}
\centering
\includegraphics{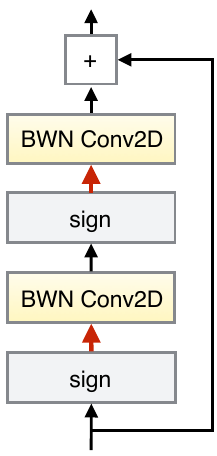}
\caption{RVAE}{(1-bit activations)}
\label{subfig:inf}
\end{subfigure}%
\begin{subfigure}{.25\textwidth}
\centering
\includegraphics{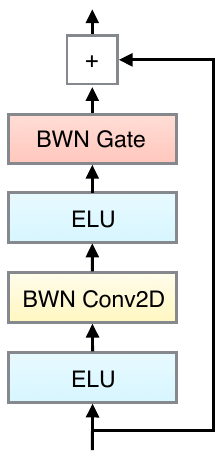}
\caption{Flow++}{(32-bit activations)}
\label{subfig:inf}
\end{subfigure}%
\begin{subfigure}{.25\textwidth}
\centering
\includegraphics{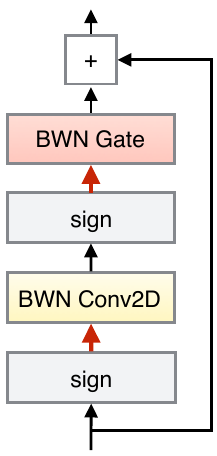}
\caption{Flow++}{(1-bit activations)}
\label{subfig:inf}
\end{subfigure}%
\caption{The residual blocks used in the binarized ResNet VAE and Flow++ models, using both binary and floating-point activations. The $\text{BWN Gate}$ layer is a binary weight normalized $1\times1$ convolution followed by a gated linear unit. We display the binary valued tensors with thick red arrows.}
\label{fig:resblocks}
\vspace{-0.5cm}
\end{figure}

\section{Deep Generative Models with Binary Weights}

We now describe the binarized versions of the ResNet VAE and Flow++ model, using the techniques and considerations from Section \ref{sec:binary_components}. Note that, for both the ResNet VAE and Flow++ models, we still retain a valid probabilistic model after binarizing the weights. In both cases, the neural networks are simply used to output distribution parameters, which define a normalized density for any set of parameter values.

\textbf{ResNet VAE.} As per Section \ref{sec:partial_bin}, we wish to binarize the residual layers of the ResNet VAE. The residual layers are constructed as convolutional residual blocks, consisting of two $3 \times 3$ convolutions and non-linearities, with a skip connection. This is shown in Figure \ref{fig:resblocks}(a)-(b). To binarize the block, we change the convolutions to BWN convolutions, as described in Section \ref{sec:norm}. We can either use real-valued activations or binary activations. Binary activations allow the network to be executed much faster, but are less expressive. We use the ELU function as the real-valued activation, and the $\text{sign}$ function as the binary activation.

\textbf{Flow++.} As with the ResNet VAE, in the Flow++ model the residual layers are structured as stacks of convolutional residual blocks. To binarize the residual blocks, we change both the $3\times3$ convolution and the gated $1\times1$ convolution in the residual block to be BWN convolutions. The residual block design is shown in Figure \ref{fig:resblocks}(c)-(d). We have the option of using real-valued or binary activations.

\section{Experiments}

We run experiments with the ResNet VAE and the Flow++ model, to demonstrate the effect of binarizing the models. We train and evaluate on the CIFAR and ImageNet $(32\times32)$ datasets. For both models we use the Adam optimizer \citep{adam}, which has been demonstrated to be effective in training binary neural networks \citep{gal_review}.

For the ResNet VAE, we decrease the number of latent variables per latent layer and increase the width of the residual channels, as compared to the original implementation. We found that increasing the ResNet blocks in the first latent layer slightly increased modelling performance. Furthermore, we chose not to model the posterior using IAF layers \cite{iaf}, since we want to keep the model class as general as possible. 

For the Flow++ model, we decrease the number of components in the mixture of logistics for each coupling layer and increase the width of the residual channels, as compared to the original implementation. For simplicity, we also remove the attention mechanism from the model, since the ablations the authors performed showed that this had only a small effect on the model performance.

Note that we do not use any techniques to try and boost the test performance of our models, such as importance sampling or using weighted averages of the model parameters. These are often used in generative modelling, but since we are trying to establish the relative performance of models with various degrees of binarization, we believe that these techniques are irrelevant.

\subsection{Density Modelling}\label{sec:main_results}

We display results in Table \ref{tab:results}. We can see that the models with binary weights and real-valued activations perform only slightly worse than those with real-valued weights, for both the ResNet VAE and the Flow++ models. For the models with binary weights, we observe better performance when using real-valued activations than with the binary activations. These results are as expected given that binary values are by definition less expressive than real values. All models with binary weights perform better than a baseline model with the residual layers set to the identity, indicating that the binary layers do learn. We display samples from the binarized models in Appendix \ref{app:samples}.

Importantly, we see that the model size is significantly smaller when using binary weights - 94\% smaller for the ResNet VAE and 90\% smaller for the Flow++ model.

These results demonstrate the fundamental trade-off that using binary layers in generative models allows. By using binary weights the size of the model can be drastically decreased, but there is a slight degradation in modelling performance. The model can then be made much faster by using binary activations as well as weights, but this decreases performance further.

\subsection{Increasing the Residual Channels}

Binary models are less expensive in terms of memory and compute. This raises the question of whether binary models could be made \textit{larger} in parameter count than the model with real-valued weights, with the aim of trying to improve performance for a fixed computational budget. We examine this by increasing the number of channels in the residual layers (from 256 to 336) of the ResNet VAE. This increases the number of binary weights by approximately 40 million, but leaves the number of real-valued weights roughly constant\footnote{There will be a slight increase, since we use real-valued weights to map to and from the residual channels.}. The results are shown in Table \ref{tab:results} and Figure \ref{fig:cifar}(c). We can see the increase the binary parameter count does have a noticeable improvement in performance. The model size increases from 13 MB to 20 MB, which is still an order of magnitude smaller than the model with real-valued weights (255 MB). It is an open question as to how much performance could be improved by increasing the size of the binary layers even further. The barrier to this approach currently is training, since we need to maintain and optimize a set of real-valued weights during training. These get prohibitively large as we increase the model size significantly.

\subsection{Ablations}

We perform ablations to verify our hypothesis from Section \ref{sec:partial_bin} that we should only binarize the residual layers of the generative models. We attempt to binarize all layers in the ResNet VAE using BWN layers, using both binary and real-valued activations. The results are shown in Figure \ref{fig:cifar}(d). As expected, the loss values attained are significantly worse than when binarizing only the residual layers. We also perform an ablation comparing the performance of BWN against batch normalization in Appendix \ref{sec:batchnorm_ablation}, demonstrating the advantages of using BWN.

\begin{table}[t]
\vspace*{-1cm}
\caption{Results for binarized ResNet VAE and Flow++ model on CIFAR and ImageNet ($32\times32$) test sets. Loss values are reported in bits per dimension. We give the percentage of the model parameters that are binary and the overall size of the model parameters. The weights and activations refer to those within the residual layers of the model, which are the targets for binarization.}
\centering
\begin{adjustbox}{width=\textwidth}
\begin{tabular}{@{}lccccccc@{}}
\toprule
                    & \multicolumn{2}{c}{\textbf{Precision}} & \multicolumn{2}{c}{\textbf{Modelling loss}} & \multicolumn{1}{l}{\textbf{\# Parameters}} & \multicolumn{1}{l}{\textbf{\% Binary}} & \multicolumn{1}{l}{\textbf{Memory cost}} \\ 
                    & Weights          & Activations         & CIFAR                                   & ImageNet ($32\times32$)                                                     &                                            &                                        &                                          \\ \midrule
\textbf{ResNet VAE} & 32-bit           & 32-bit              & 3.45                               & 4.25                                                    & 56M                                        & 0\%                                 & 255 MB                                   \\
                    & 1-bit            & 32-bit              & 3.60                             & 4.47                                                & 56M                                        & 97.1\%                                 & 13 MB                                    \\
                    & 1-bit            & 1-bit               & 3.73                               & 4.58                                                    & 56M     &   97.1\%                                 & 13 MB                                    \\ 
                    \cdashlinelr{1-8}
                    \textit{increased width} & 1-bit & 32-bit & 3.56 & - & 96M & 97.7\% & 20 MB \\
                    & 1-bit & 1-bit & 3.68 & - & 96M & 97.7\% & 20 MB \\
                    \cdashlinelr{1-8}
                    \textit{no residual} & N.A. & N.A. & 3.78 & - & 1.6M & 0\% & 6 MB \\
                    \midrule
\textbf{Flow++}     & 32-bit           & 32-bit              & 3.21                               & 4.05                                                   & 34M                                        & 0\%                                 & 129 MB                                      \\
                    & 1-bit            & 32-bit              & 3.29                               & 4.18                                              & 34M                                        & 90.1\%                                 & 14 MB                                    \\
                    & 1-bit            & 1-bit               & 3.43                               & 4.30                                                 & 34M                                        & 90.1\%                                 & 14 MB                                    \\ 
                    \cdashlinelr{1-8}
                    \textit{no residual} & N.A. & N.A. & 3.54 & - & 2.2M & 0\% & 9 MB \\
                    \bottomrule
\end{tabular}
\end{adjustbox}
\label{tab:results}
\vspace{-0.4cm}
\end{table}

\begin{figure}[t]
\centering
\makebox[\linewidth][c]{
\begin{subfigure}[b]{.51\textwidth}
\includegraphics[width=\textwidth]{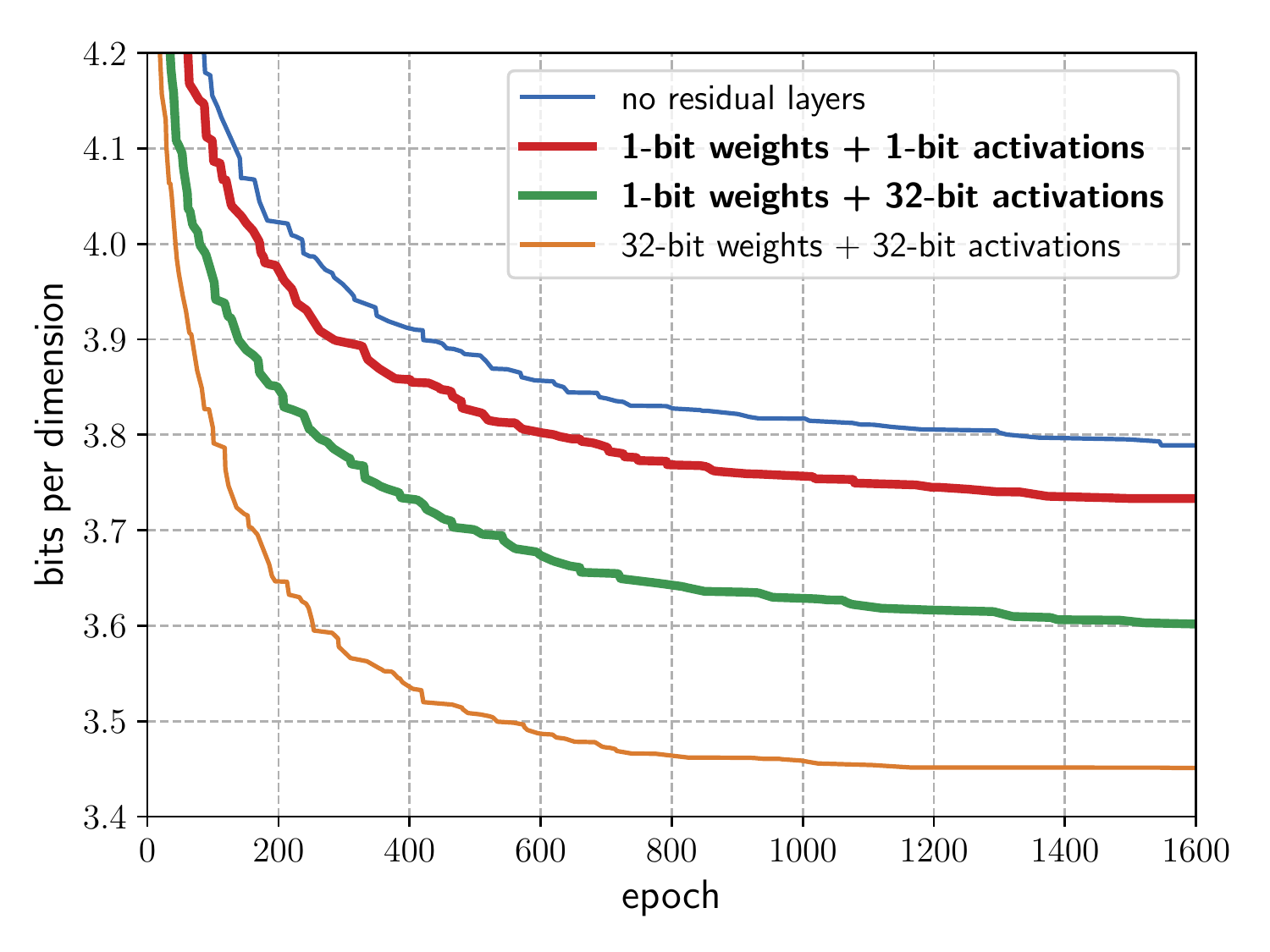}
\centering
\caption{\textbf{ResNet VAE}}
\end{subfigure}
\begin{subfigure}[b]{.51\textwidth}
\includegraphics[width=\textwidth]{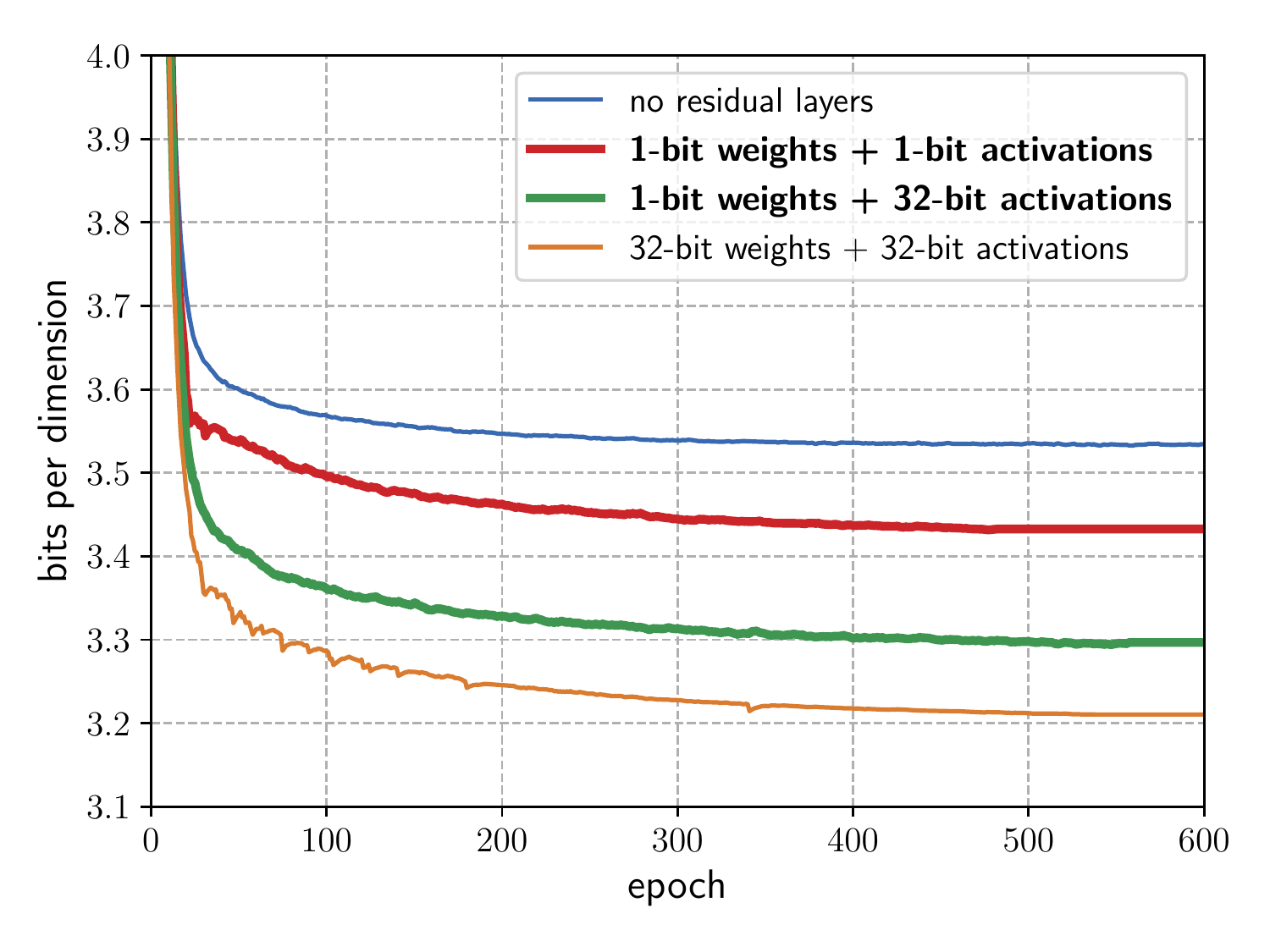}
\centering
\caption{\textbf{Flow++}}
\end{subfigure}
}
\makebox[\linewidth][c]{
\begin{subfigure}[b]{.51\textwidth}
\includegraphics[width=\textwidth]{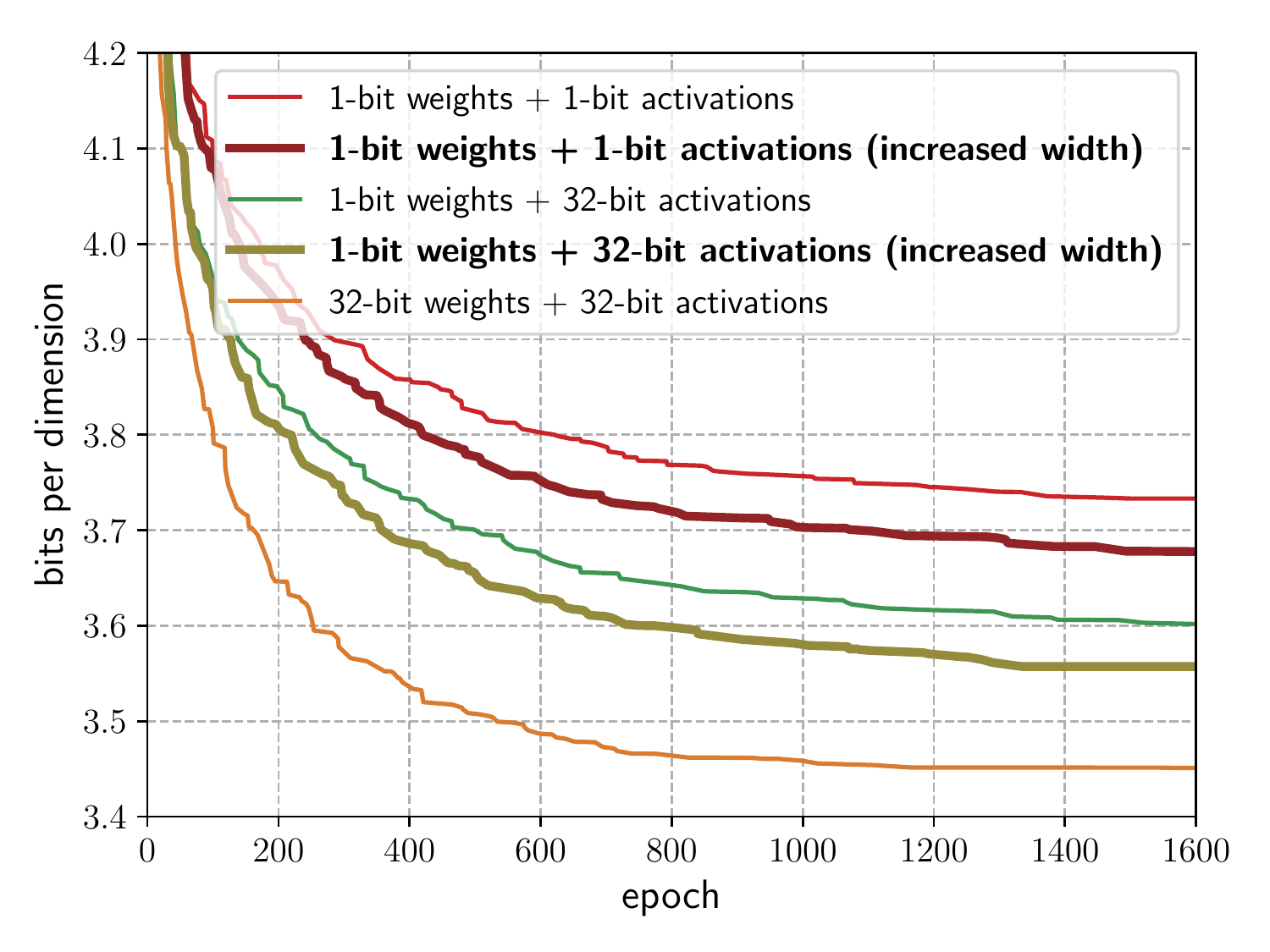}
\centering
\caption{\textbf{ResNet VAE} (increased channel width)}
\end{subfigure}
\begin{subfigure}[b]{.51\textwidth}
\includegraphics[width=\textwidth]{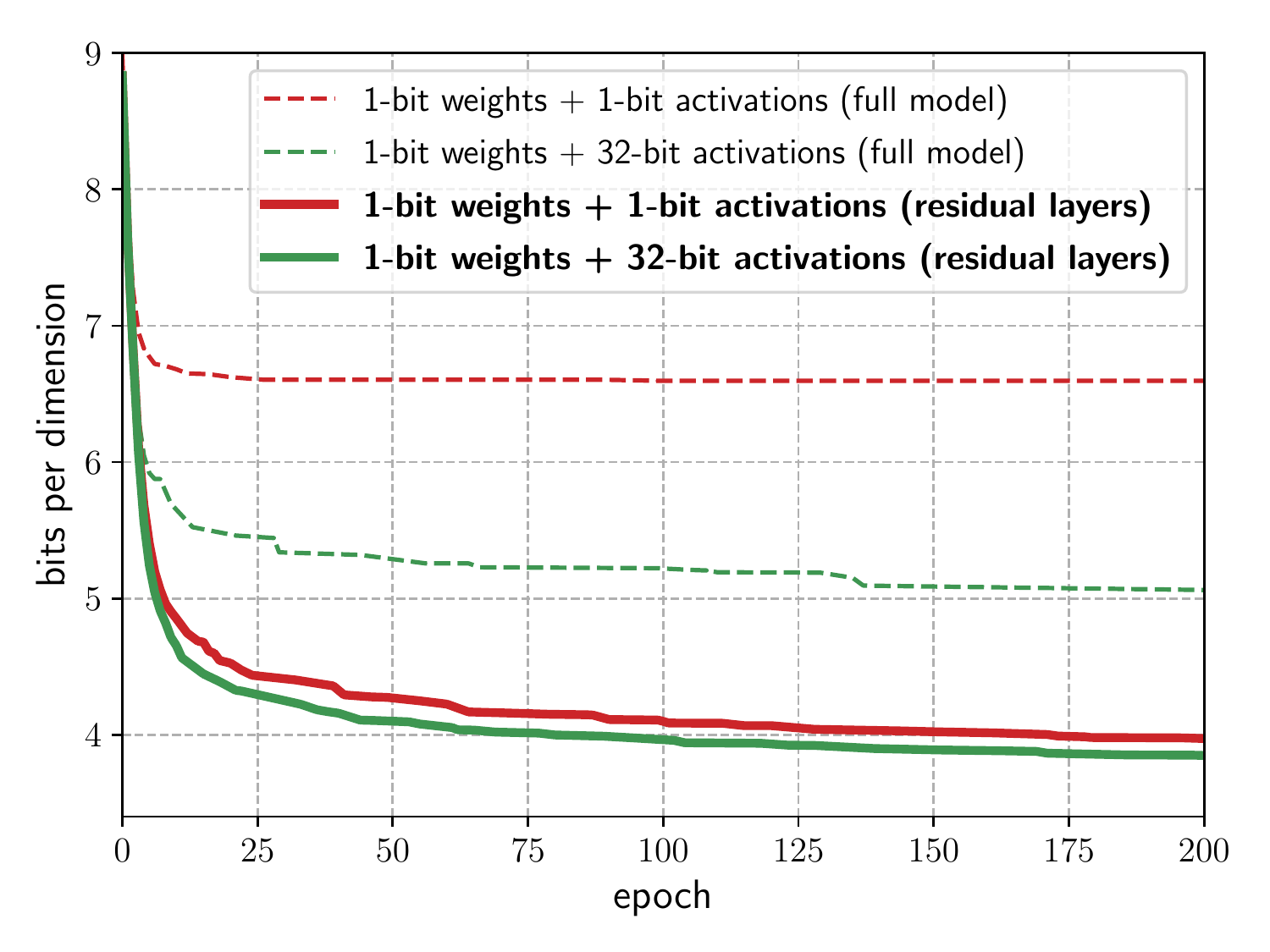}
\centering
\caption{\textbf{ResNet VAE} (ablations)}
\end{subfigure}
}
\caption{Test loss values during training of the ResNet VAE and Flow++ models on the CIFAR dataset. Subfigures (a) and (b): models with binary weights and either binary or real-valued activations. Compared to the model with real-valued weights and activations, and a baseline with the residual layers set to the identity. Subfigures (c) and (d): the effect of increasing the width of the residual channels, and ablations.}
\label{fig:cifar}
\vspace{-0.5cm}
\end{figure}

\section{Discussion and Future Work}

We have demonstrated that is possible to drastically reduce model size and compute requirements for the ResNet VAE and Flow++ models, whilst maintaining good modelling performance. We chose these models because of their demonstrated modelling power, but the methods we used to binarize them are readily applicable to other hierarchical VAEs or flow models. We believe this is a useful result for any real-world application of these generative models. such as learned lossless compression \citep{bbans, bit-swap, hilloc, lbb, idf}, which could be made practical with these reduced memory and compute benefits.



A key technical challenge that needs to be overcome for binary neural networks as a whole is the availability of implementations of the fast binary linear operations such as binary convolutions. Proof-of-concept implementations of these binary kernels have been developed \citep{bnn_bengio, xnornet, espresso}. However, there is no implementation that is readily usable as a substitute for the existing kernels in frameworks such as PyTorch and Tensorflow. 


\section{Conclusion}

We have shown that it is possible to implement state-of-the-art deep generative models using binary neural networks. We proposed using a fast binary weight normalization procedure, and shown that it is necessary to binarize only the residual layers of the model to maintain modelling performance. We demonstrated this by binarizing two state-of-the-art models, the ResNet VAE and the Flow++ model, reducing the computational cost massively. We hope this insight into the possible trade-off between modelling performance and computational cost will stimulate further research into the efficiency of deep generative models.


\newpage

\bibliography{iclr2021_conference}
\bibliographystyle{iclr2021_conference}

\newpage
\appendix

\section{Samples}\label{app:samples}

\begin{figure}[h]
\centering
\makebox[\linewidth][c]{
\begin{subfigure}[b]{.35\textwidth}
\includegraphics[width=\textwidth]{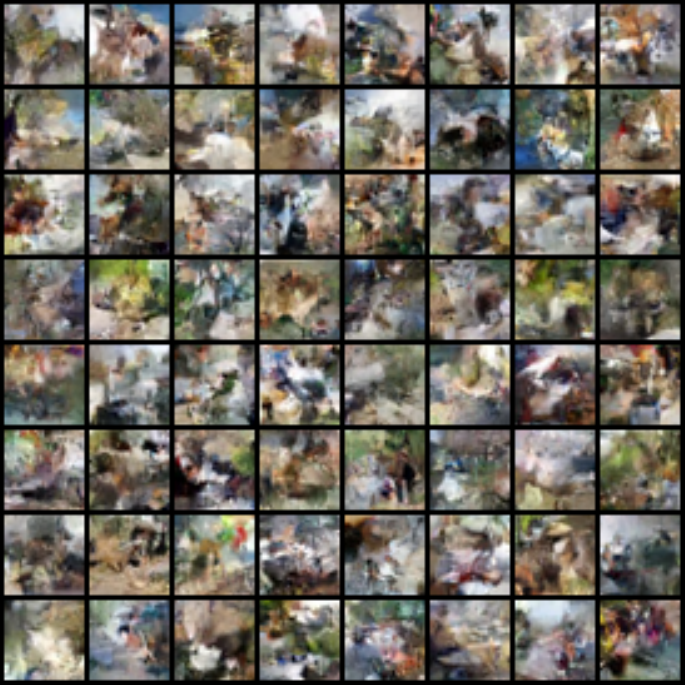}
\centering
\caption{ResNet VAE}{(32-bit weights, 32-bit activations)}
\end{subfigure}\hfill
\begin{subfigure}[b]{.35\textwidth}
\includegraphics[width=\textwidth]{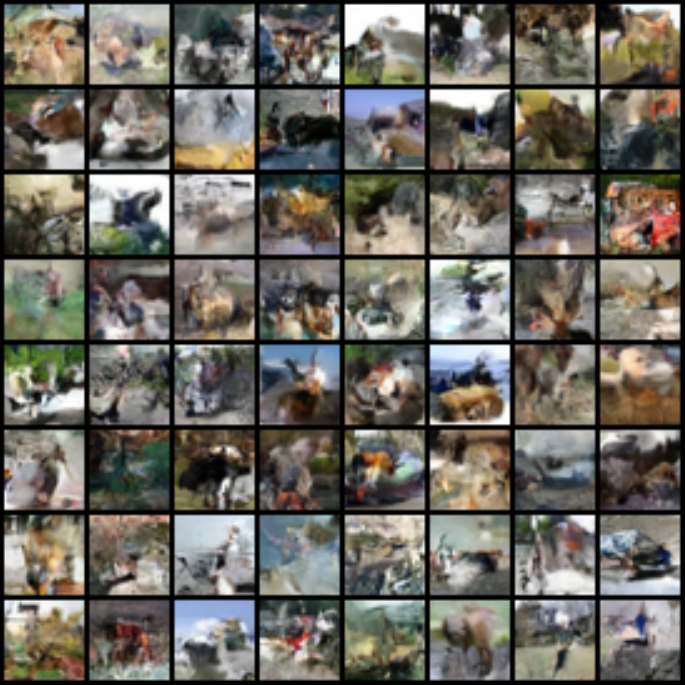}
\centering
\caption{Flow+}{(32-bit weights, 32-bit activations)}
\end{subfigure}
}

\makebox[\linewidth][c]{
\begin{subfigure}[b]{.35\textwidth}
\includegraphics[width=\textwidth]{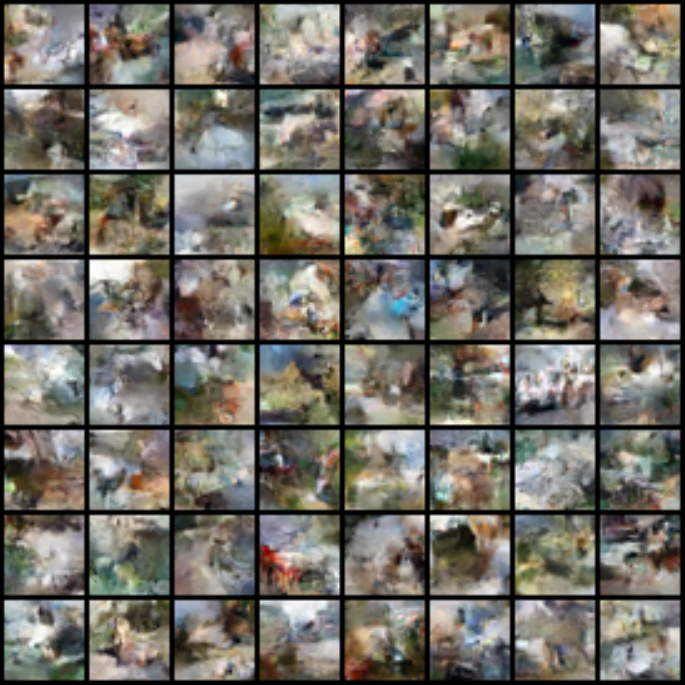}
\centering
\caption{ResNet VAE}{(1-bit weights, 32-bit activations)}
\end{subfigure}\hfill
\begin{subfigure}[b]{.35\textwidth}
\includegraphics[width=\textwidth]{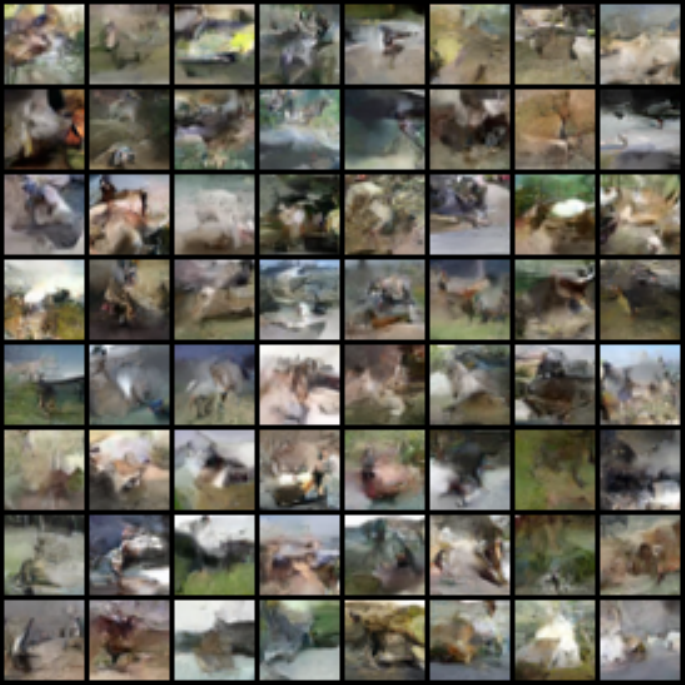}
\centering
\caption{Flow++}{(1-bit weights, 32-bit activations)}
\end{subfigure}
}

\makebox[\linewidth][c]{
\begin{subfigure}[b]{.35\textwidth}
\includegraphics[width=\textwidth]{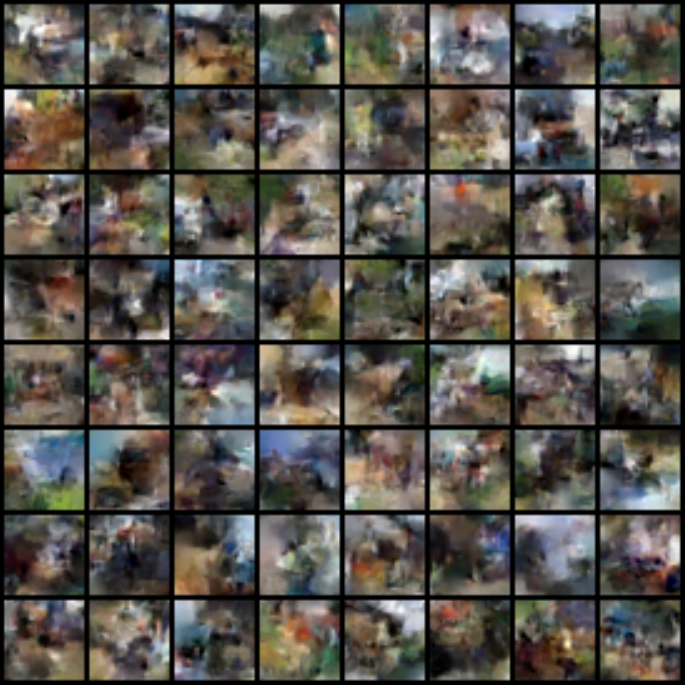}
\centering
\caption{ResNet VAE}{(1-bit weights, 1-bit activations)}
\end{subfigure}\hfill
\begin{subfigure}[b]{.35\textwidth}
\includegraphics[width=\textwidth]{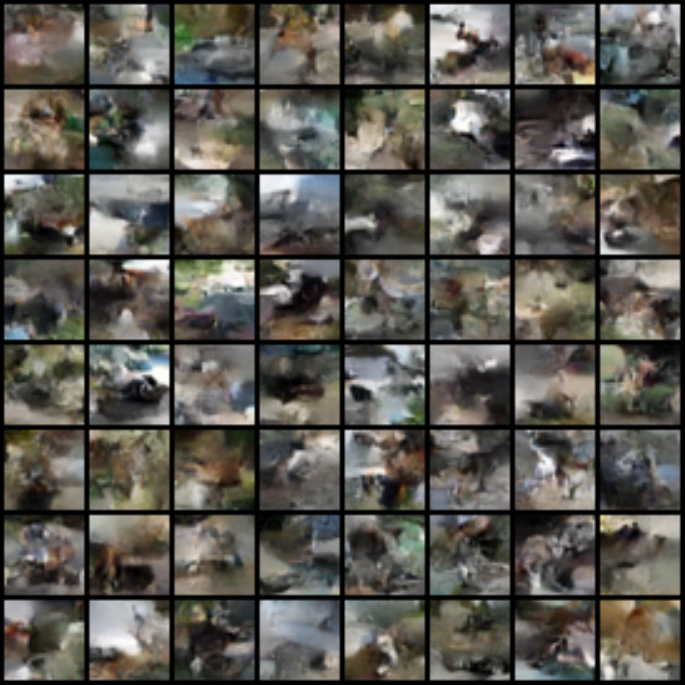}
\centering
\caption{Flow++}{(1-bit weights, 1-bit activations)}
\end{subfigure}
}
\caption{Samples from the ResNet VAE (left) and Flow++ (right) models trained on CIFAR. We provide samples from the models with (a)/(b) real-valued weights and activations, (c)/(d) binary weights and real-valued activations, (e)/(f) binary weights and activations.}
\label{fig:cifar}
\end{figure}

\newpage

\begin{figure}[t]
\centering
\makebox[\linewidth][c]{
\begin{subfigure}[b]{.51\textwidth}
\includegraphics[width=\textwidth]{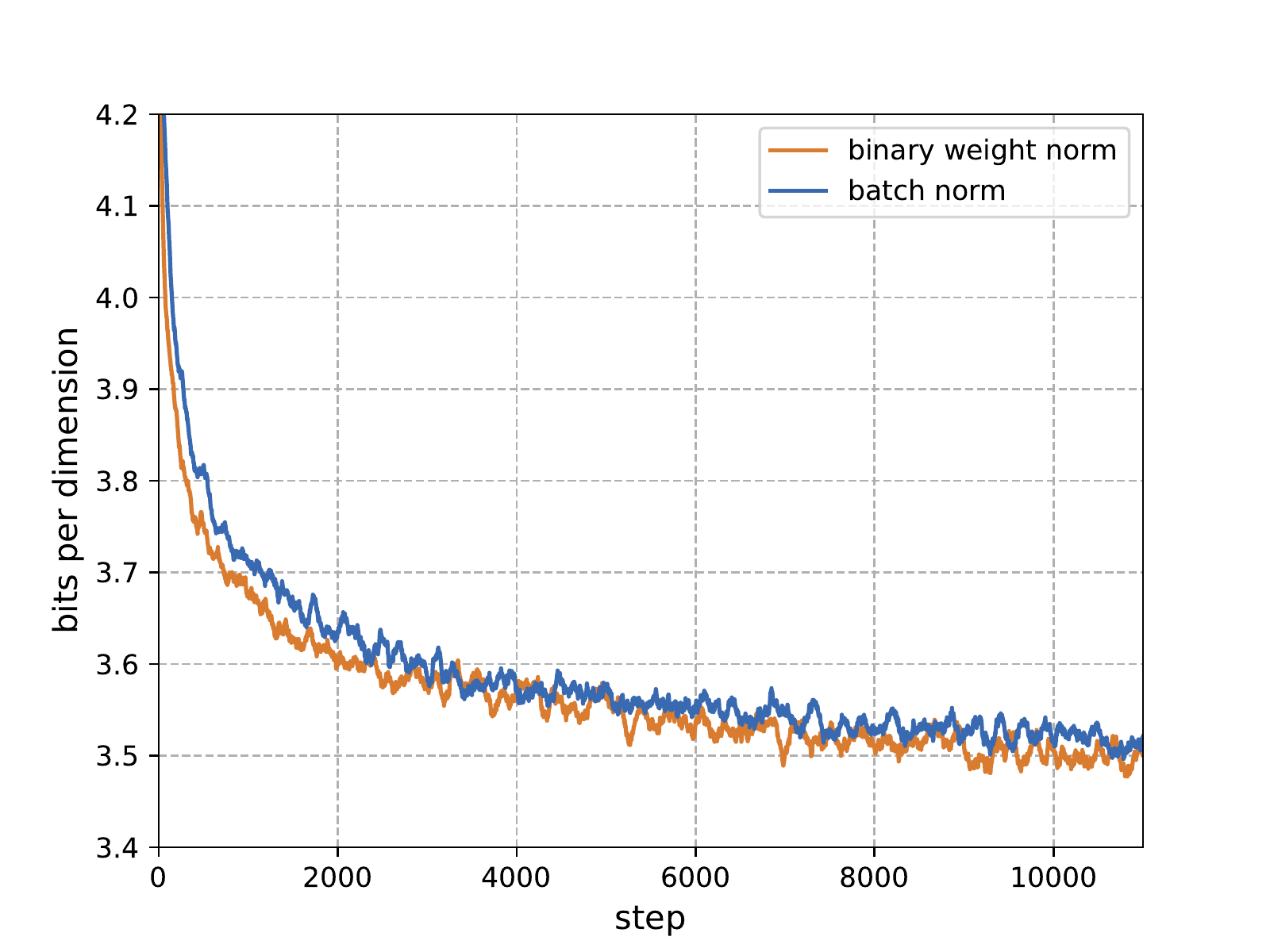}
\centering
\caption{\textbf{Flow++} (1-bit weights + 1-bit activations)}
\end{subfigure}
\begin{subfigure}[b]{.51\textwidth}
\includegraphics[width=\textwidth]{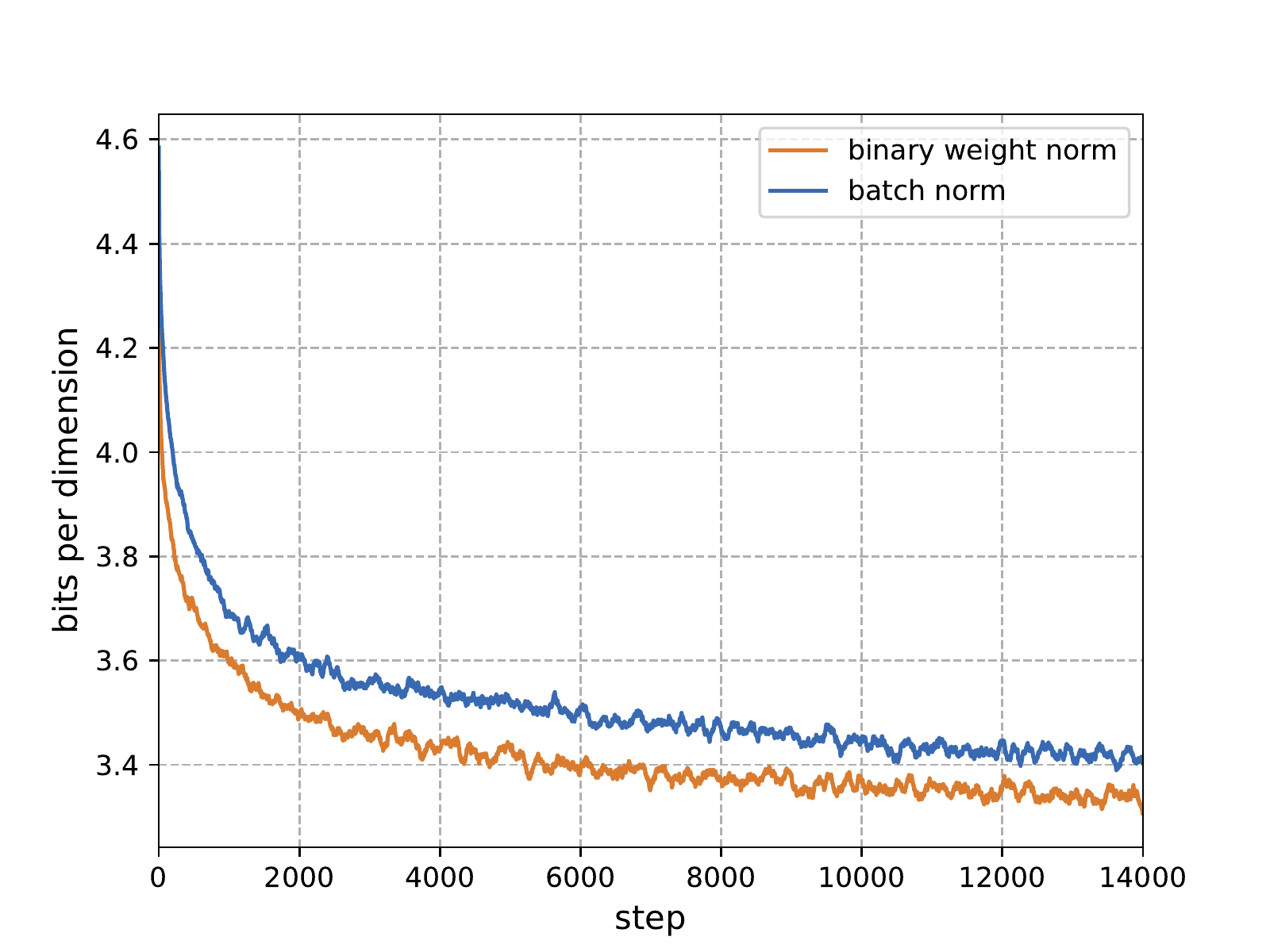}
\centering
\caption{\textbf{Flow++} (1-bit weights + 32-bit activations)}
\end{subfigure}
}
\caption{Training loss values achieved when using binary weight normalization and batch normalization for the training of binary weighted Flow++ models.}
\label{fig:batchnorm_ablation}
\end{figure}

\section{Ablation of binary weight normalization}\label{sec:batchnorm_ablation}

To examine the performance of the binary weight normalization (BWN), we perform an ablation against using the more widely used batch normalization \citep{batchnorm}. We simply place a batch normalization operation after every layer, instead of using BWN. Note that this still permits the use of fast binary operations, since the weights and activations are binary valued. We train the Flow++ model with binary weights and both binary and real-valued activations, comparing the two normalization schemes. The results are shown in Figure \ref{fig:batchnorm_ablation}. We can see that BWN is slightly better for the model with binary activations, and significantly better for the model with real-valued activations. Importantly, we have found BWN to be more stable than batch normalization, which can often result in training instabilities. Indeed, to obtain the results we present when using batch normalization, training was restarted many times. BWN is also both faster to compute and simpler, not relying on retaining running averages of batch statistics.

It is also worth noting, that it is not possible to train these binary weighted generative models without any form of normalization, since training is too unstable. This is not surprising, since the binary weights themselves are large in magnitude and can result (in particular with binary activations) in very large layer outputs.

\section{The ResNet VAE Model}\label{app:vaes}

The ResNet VAE model \citep{iaf} is a hierarchical VAE. We make some, relatively small, improvements over the original model, and now give a full description of the model. 

The model has a hierarchy of latent layers, $\bz_{1:L}$. The generative model factors as:
\begin{equation}
    \pT(\bx, \bz_{1:L}) = \pT(\bx|\bz_{1:L})\pT(\bz_L)\prod_{l=1}^{L-1}\pT(\bz_l|\bz_{l+1:L})
\end{equation}

The inference model is factored top-down:
\begin{equation}
    \qPhi(\bz_{1:L}| \bx) = \qPhi(\bz_L|\bx) \prod_{l=1}^{L-1}\qPhi(\bz_l|\bz_{l+1:L}, \bx)
\end{equation}

\begin{figure}[t]
\centering
\begin{subfigure}{.5\textwidth}
\centering
\includegraphics{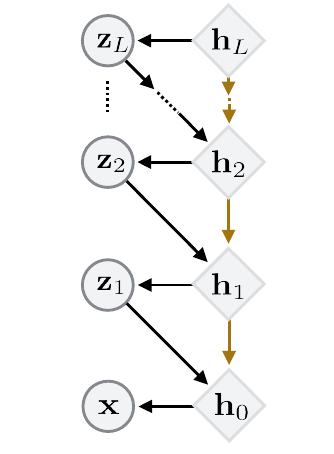}
\caption{Generative model}
\label{subfig:gen}
\end{subfigure}%
\begin{subfigure}{.5\textwidth}
\centering
\includegraphics{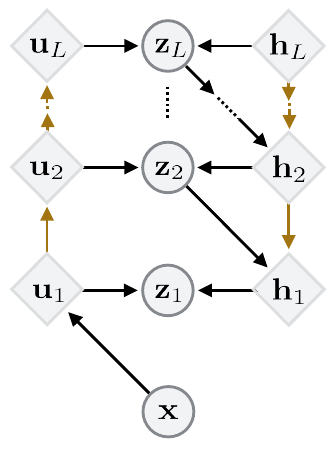}
\caption{Bi-directional inference model}
\label{subfig:inf}
\end{subfigure}%
\caption{Graphical models of the generative and inference models in a hierarchical VAE with bi-directional inference. Stochastic nodes are circular, deterministic nodes are diamond. Green lines indicate residual layers.}
\label{fig:graphicalmodel}
\end{figure}

There is also a deterministic upwards pass (through the latent layers) performed in the inference model, which produces features used by the posterior, conditioned on just $\bx$. We refer to the inference model as bidirectional, since there is both an upwards and downwards pass to be performed. The full graphical model is shown in Figure \ref{fig:graphicalmodel}.

The objective is obtained by expanding the usual ELBO:
\begin{align}
    \log p(\bx) \geq ~&\Eb{\qPhi(\bz_{1:L})}{\log \pT(\bx | \bz_{1:L})} - \KLD{\qPhi(\bz_L|\bx)}{\pT(\bz_L)} \\ &-\sum_{l=1}^{L-1}\KLD{\qPhi(\bz_l|\bz_{l+1:L}, \bx)}{\pT(\bz_l|\bz_{l+1:L})}
\end{align}
Where $D_{\text{KL}}$ is the KL divergence. Both the prior and posterior for a latent layer are diagonal when conditioned on deeper layers. We use logistic distributions for the latent variables, rather than the usual Gaussian distributions. We observed slightly improved performance using logistic distributions, and the parameterization is similar to a Gaussian, with a mean and scale parameter per dimension. The likelihood $\pT(\bx|\bz_{1:L})$ is a discretized logistic distribution.

In Figure \ref{fig:graphicalmodel} the residual connections are displayed in green, with the non-residual connections in black. The non-residual connections are convolutional layers with ELU activations functions. The residual connections are made from stacks of residual blocks. Each residual block is constructed as:
\begin{equation}
    \mathrm{Input} \rightarrow \mathrm{Activation} \rightarrow \mathrm{Conv2D_{3x3}} \rightarrow \mathrm{Activation} \rightarrow \mathrm{Conv2D_{3x3}}
\end{equation}
With a skip connection adding the output to the input. The original implementation uses just one block per layer of latents. We expand this to a stack of blocks, of varying length. This block structure is depicted for the binary case in Figure \ref{fig:resblocks}(a)-(b). For the floating-point model we simply use floating-point weight normalized convolutions, rather than BWN convolutions, and ELU activations.


\section{The Flow++ model}\label{app:flowpp}

Here we describe fully the variational dequantization from the Flow++ model \citep{flowpp}, and any alterations we make to the model itself.

As described in Section \ref{sec:gen_models}, flows are invertible functions constructed from a composition of many simpler invertible functions:
\begin{equation}
    \fT = \bbf_1 \circ \bbf_2 \circ ... \circ \bbf_L
\end{equation}
Each $\bbf_l$ is a coupling layer (\ref{eqn:flowpp_coupling}). Coupling layers are parameterized as a stack of convolutional residual blocks, with a convolution layer before and after the stack to project to and from the channel size of the residual stack. Each block is of the form:
\begin{equation}
    \mathrm{Input} \rightarrow \mathrm{Activation} \rightarrow \mathrm{Conv2D_{3x3}} \rightarrow \mathrm{Activation} \rightarrow \mathrm{Gate}
\end{equation}
Where $\mathrm{Gate}$ is a $1\times1$ convolution followed by a gated linear unit \citep{gate}. There is a skip connection adding the input to the output, along with layer normalization \citep{layernorm}. This block structure is depicted in Figure \ref{fig:resblocks} for the binary case.

Note that the original Flow++ implementation also utilizes an attention mechanism in the coupling layers, which adds significant complexity. We omit this from our model, since their ablations demonstrated that the improvement from the attention mechanism is marginal. 

In composition the coupling layers can transform a simple density to approximate the data density. The transformed density is $\pT(\bx)$, which we obtain by the change of variables formula (\ref{eqn:changeofvars}).

Our data is generally discrete, so we actually require a discrete distribution, not a continuous density. To allow this, the Flow++ model uses variational dequantization. Suppose that the data is in $[0, 1, ..., 255]^D$. We can get a discrete distribution from a continuous density by integrating over the $D$-dimensional unit hypercube:
\begin{equation}
    \PT(\bx) = \int_{[0,1)^D}\pT(\bx + \bu)d\bu
\end{equation}
Variational dequantization then proceeds by forming a lower-bound to this discrete distribution by applying Jensen's inequality:
\begin{equation}
    \log \PT(\bx) \geq \Eb{\qPhi(\bu|\bx)}{\log \pT(\bx+\bu) - \log \qPhi(\bu|\bx)}
\end{equation}
Where $\qPhi(\bu|\bx)$ is now a learned component, which "dequantizes" the discrete data. This is itself parameterized as a flow, using a composition of coupling layers as above. So our model in total consists of a main flow $\pT(\bx)$ and a dequantizing flow $\qPhi(\bu|\bx)$.


\section{Initialization of BWN Layers}\label{app:init}

An important aspect of weight normalized layers is the initialization. Since we are normalizing the weights, and not the output of a layer (like in batch normalization), at initialization a weight normalized layer has an unknown output scale. To remedy this, it is usual to use data-dependent initialization \citep{weightnorm}, in which some data points are used to set the the initial $g$ and $b$ values such that the layer output is approximately unit normal distributed.

This can be applied straightforwardly to BWN layers when training the model end-to-end, that is initializing the model at random and training til convergence. It is common, though, when training binary neural networks for classification, to use two-stage training \citep{gal_review}. This initializes the underlying weights $\bv_{\mathbb{R}}$ of binary layers with the values from a trained model with real-valued weights.

Consider what would happen if we were to try and initialize all the components of a BWN layer with those from a trained layer with real-valued weights. The $g$ and $b$ can be transferred directly, and it is logical to initialize the underlying weights $\bv_{\mathbb{R}}$ with the trained $\mathbf{v}$ values. So the magnitude of the overall weight vector $\mathbf{w}$ would remain the same in the BWN layer as in the floating-point layer, since we normalize the $\bv_{\mathbb{B}}$ vector and apply the same $g$, $b$. This initialization seems reasonable, but fails in practice. We speculate that the reason that this fails is that, although the magnitude of the weight vector remains the same after transfer, the \textit{direction} can be very different, since the sign function will change the direction of $\bv_{\mathbb{R}}$\footnote{Note that this effect is stronger in higher dimensional spaces.}. Since we perform dot products with the weight vector, the output from the initialized binary layer is very different from the trained layer.

A more considered approach is to only initialize the underlying weights $\bv_{\mathbb{R}}$ with the values from the trained network, and initialize $g$ and $b$ with data-dependent initialization as normal. This way, the data-dependent initialization can compensate for the change of direction that occurs in the binarization of $\mathbf{v}$. This method does train, but slower than initializing at random and training end-to-end. The only difference between training end-to-end and using this reduced form of two-stage training is the initial values, $\bv_0$, of the real-valued weights underlying $\bv_{\mathbb{B}}$. In the random initialization these are sampled from a Gaussian:
\begin{equation}
    \bv_0 \sim \mathcal{N}(0, 0.05)
\end{equation}
We can even normalize the trained real-valued weights such that they have the same mean and variance as the Gaussian (within a weight tensor). This still results in worse performance from the two-stage training.

\end{document}

%% file: math_commands.tex
\usepackage{amsmath,amsfonts,bm}

\def\eqref#1{equation~\ref{#1}}

\def\1{\bm{1}}

\DeclareMathAlphabet{\mathsfit}{\encodingdefault}{\sfdefault}{m}{sl}
\SetMathAlphabet{\mathsfit}{bold}{\encodingdefault}{\sfdefault}{bx}{n}

\def\gT{{\mathcal{T}}}

\newcommand{\E}{\mathbb{E}}

